\documentclass[letterpaper, 10 pt, conference]{ieeeconf}  

\IEEEoverridecommandlockouts                              

\usepackage{graphics} 
\usepackage[hyphens]{url}
\usepackage{lipsum}
\usepackage{listings}
\usepackage[font=footnotesize]{caption}
\usepackage{subcaption}
\usepackage{epsfig} 
\usepackage{times} 
\usepackage{amsmath} 
\usepackage{amsfonts}
\usepackage{cases}
\usepackage{color}
\usepackage{cases}
\usepackage{bm} 
\usepackage{hyperref}
\usepackage{algorithm}
\usepackage{algpseudocode}
\usepackage{midfloat}
\usepackage{flushend} 
\usepackage[outdir=./]{epstopdf}
\newcommand{\vect}[1]{\mathbf{#1}}
\newcommand{\mtrx}[1]{\mathbf{#1}}

\def\RR{\mathbb{R}}
\def\diag{\text{\normalfont diag}}

\title{\LARGE \bf
Autonomous task planning and situation awareness in robotic surgery*
}

\author{Michele Ginesi, Daniele Meli, Andrea Roberti, Nicola Sansonetto and Paolo Fiorini
\thanks{*This research has received funding from the European Research Council
        (ERC) under the European Union’s Horizon 2020 research and innovation
        programme, ARS (Autonomous Robotic Surgery) project, grant agreement
        No. 742671.}
\thanks{Authors are with Department of Computer Science,
        University of Verona, Strada le Grazie, 15, 37134, Verona, Italy.
        {\tt\small \{michele.ginesi, daniele.meli, andrea.roberti, nicola.sansonetto, paolo.fiorini\}@univr.it}}
}

\begin{document}

\maketitle
\thispagestyle{empty}
\pagestyle{empty}

\begin{abstract}

The use of robots in minimally invasive surgery has improved the quality of standard surgical procedures. So far, only the automation of simple surgical actions has been investigated by researchers, while the execution of structured tasks requiring reasoning on the environment and the choice among multiple actions is still managed by human surgeons. In this paper, we propose a framework to implement surgical task automation. The framework consists of a task-level reasoning module based on answer set programming, a low-level motion planning module based on dynamic movement primitives, and a situation awareness module. The logic-based reasoning module generates explainable plans and is able to recover from failure conditions, which are identified and explained by the situation awareness module interfacing to a human supervisor, for enhanced safety. Dynamic Movement Primitives allow to replicate the dexterity of surgeons and to adapt to obstacles and changes in the environment. The framework is validated on different versions of the standard surgical training peg-and-ring task.

\end{abstract}

\section{INTRODUCTION}
In the last decades, the use of robots in the operating room has provided help to surgeons in performing minimally invasive surgery, improving the precision of gestures and the recovery time for patients \cite{mack2001minimally, corcione2005advantages, vidovszky2006robotic}. At present, surgeons tele-operate slave manipulators acting on the patient using a master console. One of the main long-term goals of research in surgical robotics \cite{moustris2011evolution, camarillo2004robotic} is the development of a cognitive robotic system able to understand the scene and autonomously execute an operation, or a part of it, emulating the reasoning capabilities, and under the supervision of, a human expert. As described in \cite{yang2017medical}, increasing the level of autonomy could further improve the quality of an intervention, in terms of safety and recovery time for the patient. Moreover, it could optimize the use of the operating room, solving issues such as surgeon fatigue and reducing hospital costs. The challenges towards autonomous robotic surgery have been investigated also in Artificial Intelligence (AI) research and they include situation awareness, scene understanding to monitor and adapt the surgical workflow in real time, explainable plan generation for safety, dexterous trajectory generation, and adaptation even in small workspaces. So far, most of the research has focused on the interpretation of data from sensors to guide simple surgical actions, e.g. knot-tying \cite{chow2014novel} and drilling \cite{coulson2008autonomous}. 

In this paper, we address the problem of the automation of a simulated surgical task, where multiple actions must be coordinated and the workflow of execution is not pre-defined, but it must be determined by a cognitive system depending on the real-time information acquired by sensors. We use the \emph{da Vinci Research Kit} (DVRK) as our testbed and we focus on a more complex version of the peg-and-ring task, where both single-arm and dual-arm executions are possible, depending on dynamic conditions on rings and pegs. We propose a framework which integrates answer set programming (ASP) for task planning, and Dynamic Movement Primitives (DMPs) for trajectory generation and obstacle avoidance in real time.\\
ASP is an explainable AI tool to reason on sensory information and on prior knowledge of the task as provided from experts. DMPs allow to emulate the dexterity of surgeons learning from a small dataset of gestures. Although the examined task is still far from real surgery, to the best of our knowledge this is the first framework which combines real-time situation awareness, adaptive planning at both task and motion levels, explainable failure identification and recovery within a surgical setup. The paper is organized as follows: in Section \ref{sec:review} we review the state of the art in surgical task automation and, more generally, explainable task automation in robotics, focusing on examples of safety critical tasks. Then, in Section \ref{sec:methods} we present the framework and the task, and in Section \ref{sec:experiments} we show and discuss the experimental results. A short Section of conclusions and perspectives for future work ends the paper.

\begin{figure}[H]
    \centering
    \includegraphics[scale=0.27]{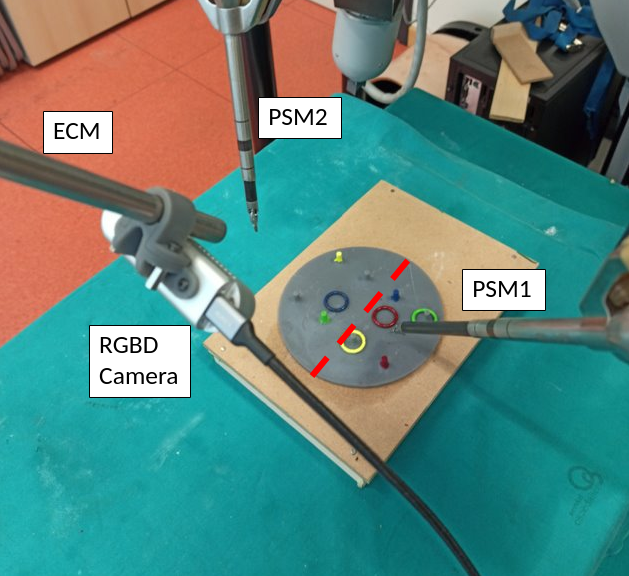}
    \caption{The setup for the peg-and-ring task. The red dashed line defines reachability regions for the two arms.} 
    \label{fig:setup}
\end{figure}

\section{STATE OF THE ART} \label{sec:review}
In recent years, several researchers have addressed the automation of basic operations in surgery. For example, in \cite{sen2016automating} the authors propose automated suturing with a mechanical needle guide to improve precision. In \cite{nagy2019dvrk} a framework for the automation of surgical sub-tasks is developed, including peg-and-ring and knot-tying tasks, integrating sensing modules and task-motion planning modules. Recently, the automation of peg-and-ring task has been proposed in \cite{minho_pegs_2020}, using RGBD camera for better accuracy. All mentioned works rely on a simple description of the tasks based on finite state machines (FSMs), assuming a static environment and neglecting situation awareness to react to anomalous events. A cognitive framework is proposed in \cite{Muradore}, where a hidden Markov model (HMM) is used to monitor the execution of needle insertion and tune stiffness parameters for the admittance control to safely drive the instrument to the goal. Statistical models as HMMs, and more general data-driven models as neural networks, are also used in robotic surgery for the interpretation of data from multiple sensors and enhance situation awareness \cite{kassahun2016surgical}. However, a huge amount of data, which is not usually available in surgery, especially for the full execution of structured tasks, is required by these techniques to achieve robust learning. Moreover, data-driven models are black-box tools, which generate plans that cannot be explained and monitored by human experts. This affects the acceptability of the autonomous system in a safety-critic scenario like surgery. Hence, we focus on knowledge-based reasoning systems, which encode prior expertise from humans and offer a clearer interpretation of the execution workflow. Knowledge representation for autonomous agents has been proposed in several robotic contexts outside surgery \cite{loutfi2008using, Tenorth}. The most popular example is Knowrob \cite{Tenorth}, where an ontology represents the general-purpose knowledge and a standard planning language \cite{Fox2003} is used to query it and defines task specifications. In \cite{ICAR19} we have proposed an ontology-based framework for the automation of the peg-and-ring task with a single industrial manipulator. However, our experiments have evidenced the limits of ontologies, which can only reason on a static prior representation of the scenario. Hence, they are not well suited when reaction to changes in the environment is required, and the knowledge base must be updated in real time. For this reason, ontologies have been mostly applied to support the monitoring of safety-critic systems, e.g. in rehabilitation \cite{DGPE12} and industry \cite{PGW12}, helping the situation understanding by human. On the contrary, non-monotonic programming offers a more flexible framework for planning \cite{dimopoulos1997encoding}, allowing reasoning on incomplete and dynamic knowledge which can be updated from sensory information. Examples in autonomous driving \cite{GOSRR18}, aerospace \cite{BGWN01} and industry \cite{FFSTT18} show the feasibility of non-monotonic reasoning in challenging and often safety-critic scenarios. The most popular tool for non-monotonic planning is Prolog \cite{ColmA90}. However, we focus on the more recent framework of Answer Set Programming (ASP) \cite{LifsV99}, which is often computationally more efficient and offers higher expressivity for task specifications, allowing preference reasoning for optimal planning \cite{BNT08}. 

\section{MATERIALS AND METHODS} \label{sec:methods}

\subsection{The peg-and-ring task}

The peg-and-ring task consists of placing rings on the same-colored pegs, using the two slave manipulators named PSM1 and PSM2 of a surgical robot such as a DVRK (see Figure \ref{fig:setup} for the setup used for the task). In order to increase the complexity of the task and show the capabilities of the reasoner, some extra specifications are added to the standard task description. First, rings may either be grasped and placed by the same arm, or they can be transferred between arms, depending on the relative position of rings and pegs with respect to the center of the base. Second, pegs can be occupied by other rings, so they must be freed before placing another ring. Finally, rings may not be visible at the beginning, or they can either be on pegs or on the base, thus requiring extraction. The positions of rings can change in real time, so the plan must be continuously adapted to the current environment, and the motion trajectories must reach moving goals.

Even if peg-and-ring is not a proper surgical task, it is widely used as a training exercise for surgeons, since it presents several challenges in common with real surgery, and therefore we see it as a necessary first step before addressing more realistic tasks. In fact, the slave manipulators must move in a surgical-scale environment, avoiding obstacles (e.g., parts of the anatomy), grasping and positioning small objects with precision and dexterity (like needle grasping in suturing). Moreover, the scene can change in real time as in a real patient's anatomy. Then, the system must be able to fast re-plan in response to new conditions and failures in an explainable way, guaranteeing the satisfaction of the final goal.

\subsection{The framework}
\begin{figure}
    \centering
    \includegraphics[scale=0.25]{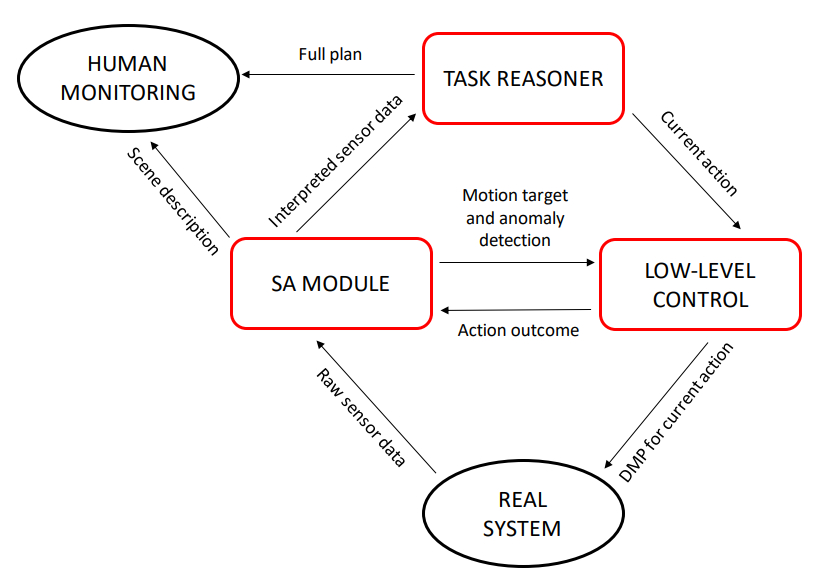}
    \caption{The proposed framework for surgical task automation. Functional modules of the framework are highlighted in red, while arrows show the stream of information between modules, the real system and an external human observer.}
    \label{fig:framework}
\end{figure}

A scheme of our proposed framework is shown in Figure \ref{fig:framework}. The exchange of information between its modules and the real system (robot + sensors) is shown. Moreover, the flow of information towards an external human observer is represented, in order to emphasize the aim for explicability and situation awareness. A description of the functions of each module and details about their integration follows.

{}

\subsubsection{Task reasoning module.}
A task reasoner based on ASP is implemented in this module. An answer set program defines the entities and specifications of the task, in terms of Boolean variables named \textit{atoms} and logical implications on atoms, named \textit{rules}. Entities are the objects involved and the actions, while specifications define rules that describe the effects and pre-conditions of actions, task constraints and the goal.

For the peg-and-ring task, we define entities as the agents \texttt{Arm} (PSM1 and PSM2), the \texttt{ring} and the \texttt{peg} with their \texttt{Color} (red, green, blue, yellow and grey). Actions with pre-conditions and effects are defined as follows:
\begin{itemize}
    \item \texttt{move(Arm, ring, Color)} to move to a colored ring, with pre-condition \texttt{reachable(Arm, ring, Color)} and effect \texttt{at(Arm, ring, Color)};
    \item \texttt{move(Arm, peg, Color)} to move to a colored peg, with pre-condition \texttt{reachable(Arm, peg, Color)} and effect \texttt{at(Arm, peg, Color)};
    \item \texttt{move(Arm, center)} to move to the transfer point, with pre-condition \texttt{in\_hand(Arm, ring, Color)} and effect \texttt{at(Arm, center)};
    \item \texttt{grasp(Arm, ring, Color)} to grasp a colored ring, with pre-condition \texttt{at(Arm, ring, Color)} and effect \texttt{in\_hand(Arm, ring, Color)};
    \item \texttt{release(Arm)} to open the gripper, with pre-condition \texttt{closed\_gripper(Arm)} and effect \texttt{not in\_hand(Arm, ring, Color)};
    \item \texttt{extract(Robot, ring, Color)} to remove a colored ring from a peg, with pre-condition \texttt{in\_hand(Arm, ring, Color)} and effect \texttt{not on(ring, Color1, peg, Color2)}.
\end{itemize}
The atom \texttt{reachable} states that a ring or a peg can be reached by an arm, depending on its relative position with respect to the center of the base.
Some atoms are defined as \textit{external}, namely they can be set by other programs, in order to allow integration with sensors. External atoms are \texttt{reachable}, \texttt{on}, \texttt{closed\_gripper} and \texttt{in\_hand}. Additionally, the atom \texttt{distance(Arm, ring, Color, Value)} is introduced to define the distance between rings and arms. This will be used for optimal plan generation in Section \ref{sec:experiments}, exploiting pre-defined constructs for preference reasoning and optimization in ASP. External atoms are received by the situation awareness module when changes in the environment are detected.
We also define executability constraints to implement user-defined specifications: 
\begin{itemize}
    \item a ring cannot be grasped by an arm with closed gripper;
    \item a ring which is on a peg cannot be moved before extraction;
    \item a ring cannot be placed on an occupied peg.
\end{itemize}
These types of constraint can be interpreted as safety requirements in real surgery.
The goal is defined as the constraint that all reachable rings are placed on their pegs, assuming \texttt{on(ring, Color, peg, Color)} is an effect of \texttt{at(Arm, peg, Color), in\_hand(Arm, ring, Color), release(Arm)}.

Given this task description, the ASP Solving Algorithm (\ref{alg:ASP}) based on SAT solving \cite{nieuwenhuis2006solving} is executed. First, \emph{grounding} of the external atoms as received from sensors is performed. This assigns an initial truth value to the corresponding Boolean variables. Then, the solver checks the holding pre-conditions and matches them with the effects of possible actions, incrementing a discrete time step until the goal is satisfied. We assume one-step delay between pre-conditions, actions and effects. Finally, the sequence of actions which minimizes the time horizon to reach the goal is returned. 
It is relevant to notice that the specifications do not determine a fixed temporal sequence of the actions as in standard FSMs, but they only define high-level task-related knowledge which must be taken into account by the ASP solver to produce the fastest feasible plan. 
By default, we use the \emph{aggregate} construct (\texttt{0 \{ Action: Pre-condition \} 1}) from ASP to force the solver to return at most one action per time step. However, in the experimental section we will relax this constraint to \texttt{0 \{ Action: Pre-condition \} 1 :- arm(Arm)}, which allows one action \emph{per robot} at each time step. Therefore, the reasoner will automatically decide whether the arms should co-operate or act independently, reducing the time to conclude the task with respect to a standard human execution.

\begin{algorithm}
    \caption{ASP Solving Algorithm}\label{alg:ASP}
    \begin{algorithmic}[1]
        \State \textbf{Input}: ASP program with specifications, external atoms
        \State \textbf{Output}: Plan
        \State \textit{Ground external atoms}
        \State \textit{Plan = [], t = 1, Action = null}
        \While{not goal}
            \If{Action != null}
                \State \textit{Ground effects of Action}
            \EndIf
            \State \textit{Check pre-conditions for actions at t}
            \If{some actions are possible}
                \State \textit{Select Action with effect closest to goal}
                \State \textit{Plan.append(Action(t))}
                \State \textit{t++}
            \Else \  \Return{Unsatisfiable}
            \EndIf
        \EndWhile
        \State \Return{Plan}
    \end{algorithmic}
\end{algorithm}

\subsubsection{Low-level control module}
This module receives the actions computed by the task reasoner, and executes the corresponding low-level control policies in temporal sequence. For each \texttt{move} action, a trajectory for the specified arm and target is computed using DMPs \cite{HPPS09}.

DMPs consist of a system of second order ODEs (one equation for each dimension of the ambient space) with a perturbation term.
The aim of DMPs is to model the perturbation term in such a way to be able to generalize the trajectory to new start and goal positions, while maintaining the shape of the learned trajectory.
The $d$-dimensional formulation is given by
\begin{subnumcases}{\label{eqs:dmp_vect}}
    \tau \dot{\vect{v}} = \mtrx{K} (\vect{g} - \vect{x}) - \mtrx{D} \vect{v} - \mtrx{K} (\vect{g}- \vect{x}_0)s + \mtrx{K} \vect{f}(s) \label{eq:new_dmp_vect_acc}\\
    \tau \dot{\vect{x}} = \vect{v}.
\end{subnumcases}
where $ \vect{x}, \vect{v} \in \RR^d $ are, respectively, position and velocity of a point (end-effector) of the system.
Matrices $ \mtrx{K}, \mtrx{D} \in \RR_+^{d \times d} $ are diagonal (\(\mtrx{K} = \diag(K_1, K_2, \ldots, K_d)\), \(\mtrx{D} = \diag(D_1, D_2, \ldots, D_d)\)) and satisfy the \emph{critical damping condition} \( D_i = 2\sqrt{K_i} \).
$ \vect{g}, \vect{x}_0 \in \RR^d $ are, respectively, the goal and starting position, and $\tau \in \RR_+$ is a time-scaling parameter.
Function $ \vect{f}:\RR \to \RR^d $ is the perturbation term.
$ s\in\RR_+ $ is a re-parametrization of time governed by the so-called \emph{canonical system},
\[
    \tau \dot{s} = -\alpha s,\quad \alpha \in \RR_+,
\]
with initial condition $ s(0) = 1 $.

During the learning phase, a desired trajectory $ \widetilde{\vect{x}}(t) \in \RR^d , \, t\in[0, T]$ is recorded.
By fixing the elastic and damping parameters $K_i, D_i, i=1,2,\ldots,d$, and imposing $\tau = 1$ and $ \vect{x}_0 = \widetilde{\vect{x}}(0) $, $ \vect{g} = \widetilde{\vect{x}}(T) $, we can solve \eqref{eq:new_dmp_vect_acc} to compute the \emph{desired forcing term} $\widetilde{\vect{f}}(s(t))$.
Next, we approximate $\widetilde{\vect{f}}(s)$ in each dimension using basis functions $ \psi_i(s) $:
\begin{equation*}
    \widetilde{f}(s) \approx f(s) =  \frac{\sum_{i=0}^N \omega_i\,\psi_i(s)}{\sum_{i=0}^N \psi_i(s)} \, s.
\end{equation*}
We use, as set of basis functions $\psi_i$, the \emph{mollifier-like basis functions} proposed in \cite{ginesi2019dmp}.
The weights $\omega_i$ are computed so as to minimize the $L_2$-error between the desired and the approximated forcing term \( \| \widetilde{f} - f \|_2 \).

Once the weights $\omega_i$ have been computed, \eqref{eqs:dmp_vect} can be solved changing $\vect{x}_0$, and $ \vect{g} $. The set of weights actually defines the low-level policy for the specific action to be executed.
Moreover, by changing $\tau$ it is possible to change the speed of execution of the trajectory.
In order to make the execution more robust against changes of starting and goal position, we use the approach presented in \cite{ginesi2019dmp} to make DMPs invariant under rotation and dilatation of the relative position $\vect{g} - \vect{x}_0$.

In order to apply the DMPs framework to our task, we need to model also the orientation of the end-effector to replicate the dexterity of surgeons.
To do so, we rely on the DMPs formulation in unit quaternion space presented in \cite{SFL19}.
We emphasize that the same canonical system is `shared' among Cartesian and orientation DMPs, so that all the trajectories are synchronized.

Obstacle avoidance is implemented in the DMP framework by using the method proposed in \cite{GMCDSF19}, in which an obstacle is modeled as a repulsive potential field, whose negative gradient is added to \eqref{eq:new_dmp_vect_acc} to perturb the trajectory. In our scenario, obstacles are represented by pegs.

The execution of a DMP can be interrupted if an anomaly is detected by the situation awareness module, in which case a new plan generation is requested.

\subsubsection{Situation awareness (SA) module}
This module is in charge of the semantic interpretation of data from sensors, providing a high-level description of the environment in real time. This allows non-expert users to understand and monitor the worflow of execution, enhancing explainability and safety of the framework. Moreover, the SA module acts as an intermediate layer between task- and motion-level modules, improving the scalability and generality of the framework. 
The inputs to the SA module are the real-time poses of pegs and centers of rings from a RGBD camera, and the poses of the arms from kinematics.
These poses are computed with respect to a common frame \emph{world} for both the camera and the robotic arms using hand-eye calibration.

During the execution of the task, the Vision Algorithm (\ref{alg:VA}) subsamples the point cloud from the scene in order to guarantee real-time performances. The base and the pegs are assumed to be static during the whole execution, and they are identified only at the beginning of the task. Then, the poses of all rings are retrieved at each time step. The identification of pegs and rings is performed in two steps. First, color segmentation allows to identify same-colored points. Then, Euclidean clustering allows to separate the clouds of ring and peg. Finally, Random Sample Consensus (RANSAC) is used to fit a torus shape on both clusters, and the best fitting cluster is identified as the ring, while the other is identified as the peg. 

\begin{algorithm}
    \caption{Vision Algorithm}\label{alg:VA}
    \begin{algorithmic}[1]
        \State \textbf{Input}: Point Cloud $P_{in}$ in real time 
        \State \textbf{Output}: Poses of rings $Pose_{ring}$ and pegs $Pose_{peg}$
        \For{ t = 1 to $\infty$ }
            \State \textit{Subsample $P_{in}(t)$ to $P_{sub}(t)$}
            \If {t = 1}
                \State \textit{Plane estimation}
                \State \Return{$Pose_{peg}$}
            \Else 
                \For{ colorID = 0 to 3 } 
                    \State \textit{Color Segmentation of $P_{sub}(t)$ }
                    \State \textit{Euclidean clustering}
                    \State \textit{Ring identification $\gets$ RANSAC }
                    \State \Return {$ Pose_{ring}(t)[colorID]$ }
                \EndFor
            \EndIf
        \EndFor
        
    \end{algorithmic}
\end{algorithm}

The output of the vision algorithm is used by the SA Algorithm (\ref{alg:SA}) to provide external atoms to the task reasoner, check failure conditions and compute targets for the low-level control module.
When an action is started by the low-level controller, the specific failure conditions and target pose are computed during the whole execution.
In detail, when moving to a ring, the grasping point is selected as the point of the ring cloud which is the most distant from the pegs. Given the position $r$ of a generic point on the ring and the set $X$ of positions of pegs, the function to be maximized is chosen as $\sqrt{\sum_{p \in X} \ \left\lVert r-p \right\rVert _2^2}$. In this way, we guarantee collision-free grasping. Given the grasping point, the orientation is chosen such that the gripper approaches the ring orthogonally to the ring's plane. 
When moving to a peg, the target orientation is chosen to be orthogonal to the plane, so that the DMP can automatically recover in case the ring flips. Finally, when transferring between the two arms occurs, the target point for the free arm is chosen as the one opposite to the grasping point of the main arm.

In case an anomaly is identified, the low-level control module is notified and the updated external atoms are sent to the task reasoner to compute a new plan.

\begin{algorithm}
    \caption{SA Algorithm}\label{alg:SA}
    \begin{algorithmic}[1]
        \State \textbf{Input}: Action, $Pose_{ring}$, $Pose_{peg}$
        \State \textbf{Output}: Failure message, target pose, external atoms
        \State \textit{failure = True}
        \If{failure}
            \State \textit{Compute\_externals($Pose_{ring}$, $Pose_{peg}$)}
            \State \Return{external atoms}
            \State \textit{failure = False}
        \ElsIf{Executing Action}
            \While{Action not ended}
                \State \textit{Compute\_target($Pose_{peg}$)}
                \State \Return{target pose}
                \If{Action = \texttt{move\_ring}}
                    \If{$Pose_{ring}[colorID]$ is not retrieved}
                        \State \textit{failure = True}
                        \State \Return{failure}
                    \EndIf
                \ElsIf{Action = \texttt{move\_peg}}
                    \If{ring fallen or peg occupied}
                        \State \textit{failure = True}
                        \State \Return{failure}
                    \EndIf
                \ElsIf{Action = \texttt{move\_center}}
                    \If{ring fallen}
                        \State \textit{failure = True}
                        \State \Return{failure}
                    \EndIf
                \EndIf
            \EndWhile
        \EndIf
    \end{algorithmic}
\end{algorithm}

\section{EXPERIMENTAL RESULTS} \label{sec:experiments}
\begin{figure}[H]
\centering
\begin{subfigure}{0.23\textwidth}
\includegraphics[width=\textwidth]{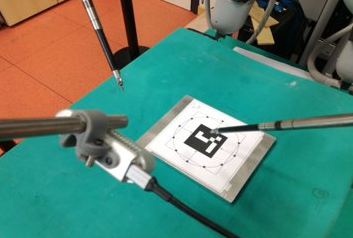}
\caption{\label{fig:calib}}
\end{subfigure}
\hfill
\begin{subfigure}{0.23\textwidth}
\includegraphics[width=\textwidth]{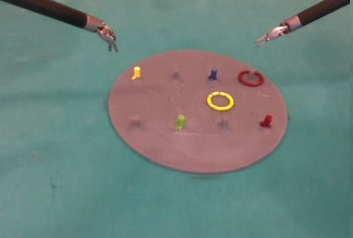}
\caption{\label{fig:test_A}}
\end{subfigure}
\begin{subfigure}{0.23\textwidth}
\includegraphics[width=\textwidth]{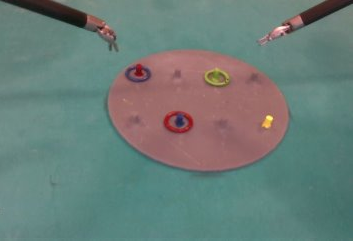}
\caption{\label{fig:test_B}}
\end{subfigure}
\hfill
\begin{subfigure}{0.23\textwidth}
\includegraphics[width=\textwidth]{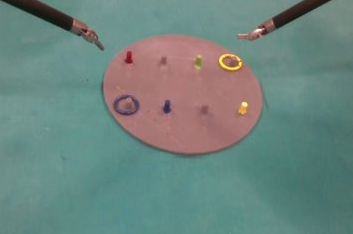}
\caption{\label{fig:test_C}}
\end{subfigure}
\caption{Custom calibration board (top-left) and the tested scenarios as seen from the Realsense.}
\label{fig:tests}
\end{figure}

For the experimental evaluation of the framework, we use our da Vinci Research Kit (DVRK). The communication between modules of the framework relies on ROS infrastracture. The task reasoning module is implemented using the state-of-the art grounder and solver Clingo \cite{gebser2008user}, which offers Python APIs for easy integration with ROS, as well as useful tools for incremental time-horizon solving, definition of external atoms and optimization statements. 
We use an Intel RealSense d435 camera, which allows us to see depth images from at least $0.105$m and obtain the point cloud of the scene. The Point Cloud Library (PCL) is used to process the stream from the camera, since it offers a standard easy integration with ROS, as well as useful tools for RANSAC segmentation. The hand-eye calibration is performed using a custom calibration board with a marker on the top (Figure \ref{fig:calib}), following the procedure described in \cite{Andrea}. The RGBD camera is rigidly attached to the end-effector of the endoscopic arm of the da Vinci (ECM), with a properly designed adapter. Our calibration procedure allows to reach an accuracy of 1 mm in pose detection, needed for the small size of the setup.

\begin{figure*}[t]
    \centering
    \begin{subfigure}{0.45\textwidth}
        \includegraphics[width=1\linewidth]{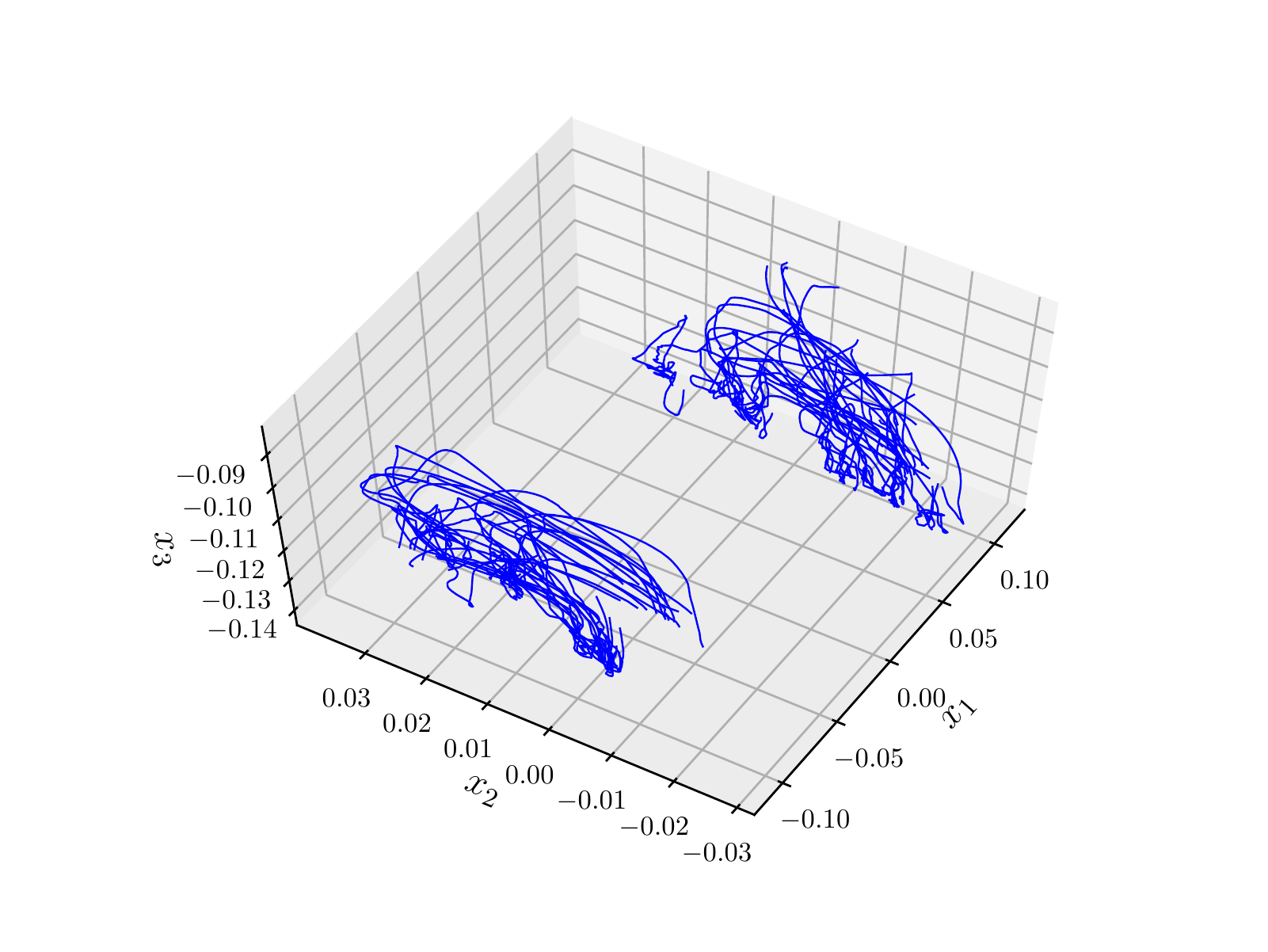}
        \caption{\label{subfig:trj_set}}
    \end{subfigure}
    \begin{subfigure}{0.45\textwidth}
        \includegraphics[width=1\linewidth]{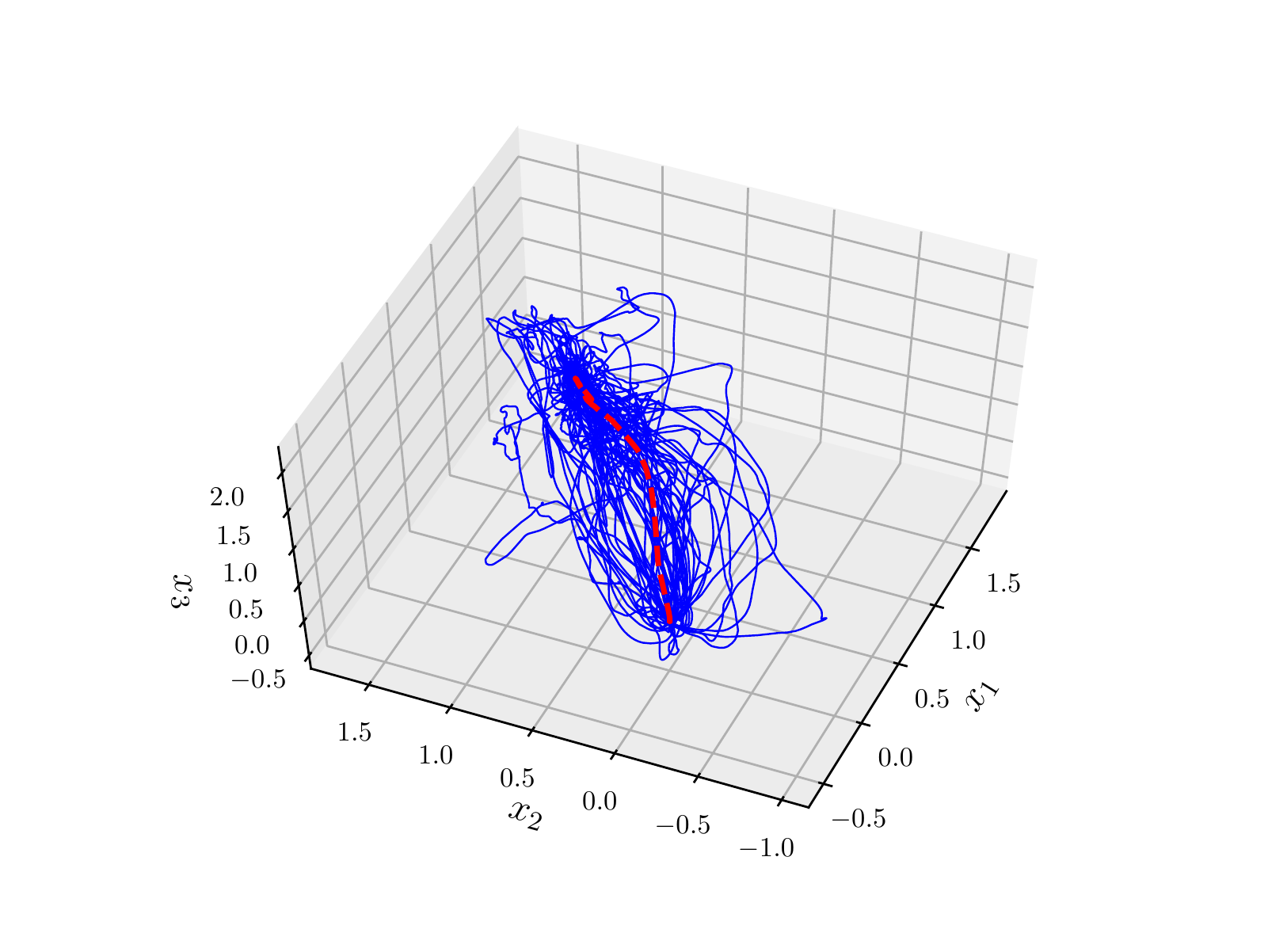}
        \caption{\label{subfig:trj_gen}}
    \end{subfigure}
    \caption{a) The set of Cartesian trajectories for the \texttt{move\_{ring}} gesture, for both PSMs; b)
    trajectories after roto-dilatation, to start at the origin, $ \mathbf{x}_0 = \mathbf{0} $ and end at the vector of ones, $ \vect{g} = \vect{1} $ for batch learning. The learned DMP is shown in red dashed line.}
    \label{fig:learning}
\end{figure*}

The DMPs for \texttt{move} actions are learned from multiple human executions using the approach presented in \cite{ginesi2019dmp}.
Three users with different dexterity performed five trials each of the peg-and-ring task tele-operating the slave manipulators.
In each execution the initial position of the rings and the pegs is kept the same, as well as the order of the rings (red, green, blue, yellow). Rings are always transferred between the two arms. 
Hence, we get 120 executions of the \texttt{move(Arm, ring, Color)} gesture (at the beginning and during transfer for each ring), 60 executions of the \texttt{move(Arm, peg, Color)} and \texttt{move(Arm, center)} gestures. Figure \ref{fig:learning} shows the learned Cartesian DMP for the \texttt{move(Arm, ring, Color)} gesture as an example.

\begin{figure*}[t]
\centering
\begin{subfigure}{0.22\textwidth}
\includegraphics[width=\textwidth]{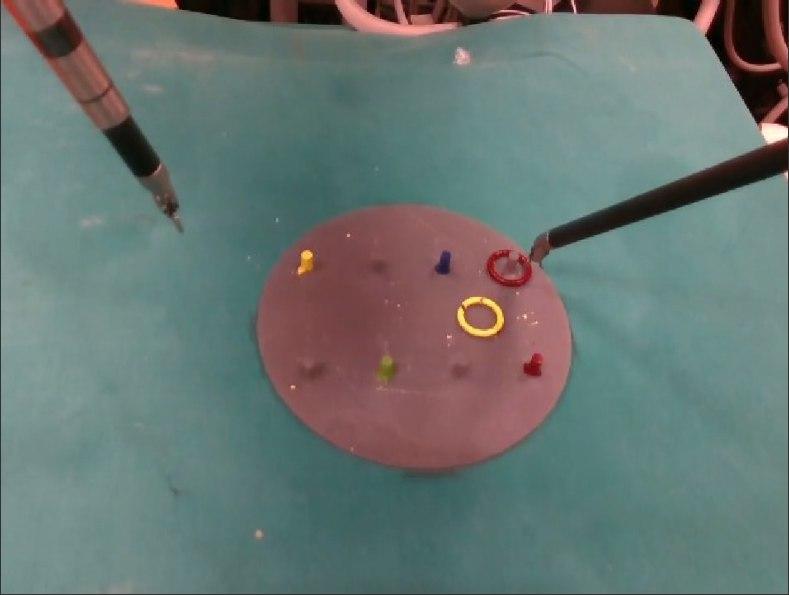}
\caption{move(psm1,ring,red,1)\label{fig:exec1}}
\end{subfigure}
\begin{subfigure}{0.22\textwidth}
\includegraphics[width=\textwidth]{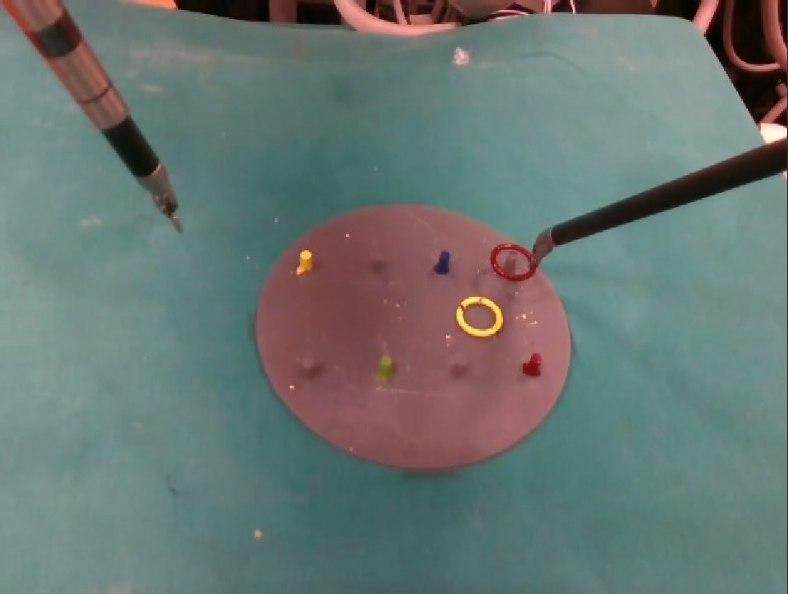}
\caption{extract(psm1,ring,red,3)\label{fig:exec2}}
\end{subfigure}
\begin{subfigure}{0.22\textwidth}
\includegraphics[width=\textwidth]{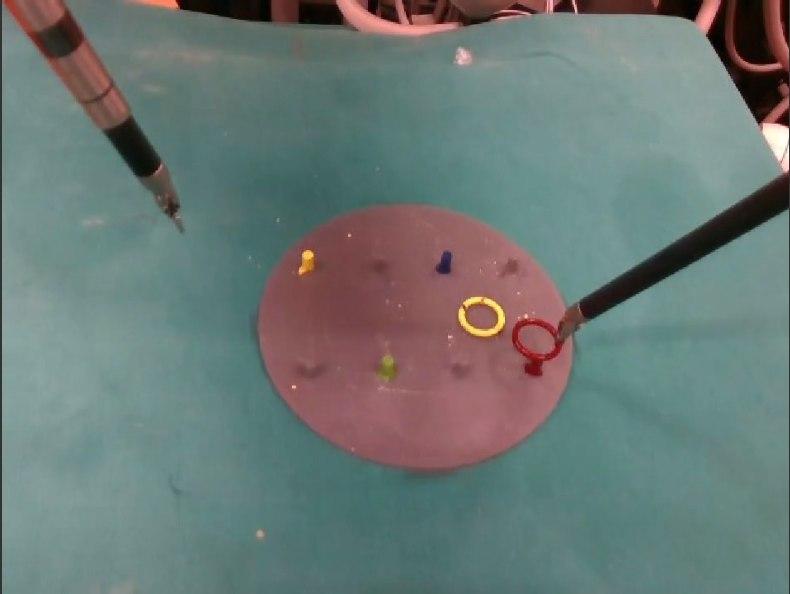}
\caption{move(psm1,peg,red,4)\label{fig:exec3}}
\end{subfigure}
\begin{subfigure}{0.22\textwidth}
\includegraphics[width=\textwidth]{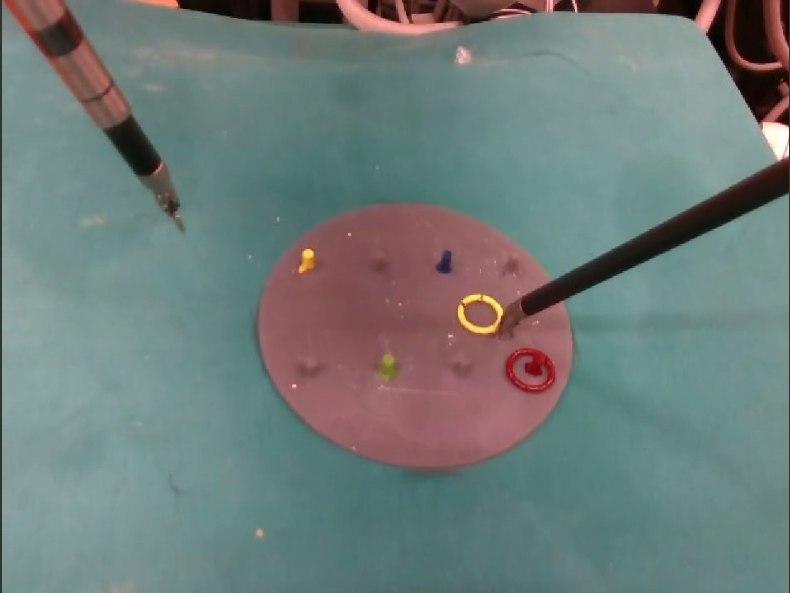}
\caption{move(psm1,ring,yellow,6)\label{fig:exec4}}
\end{subfigure}
\begin{subfigure}{0.22\textwidth}
\includegraphics[width=\textwidth]{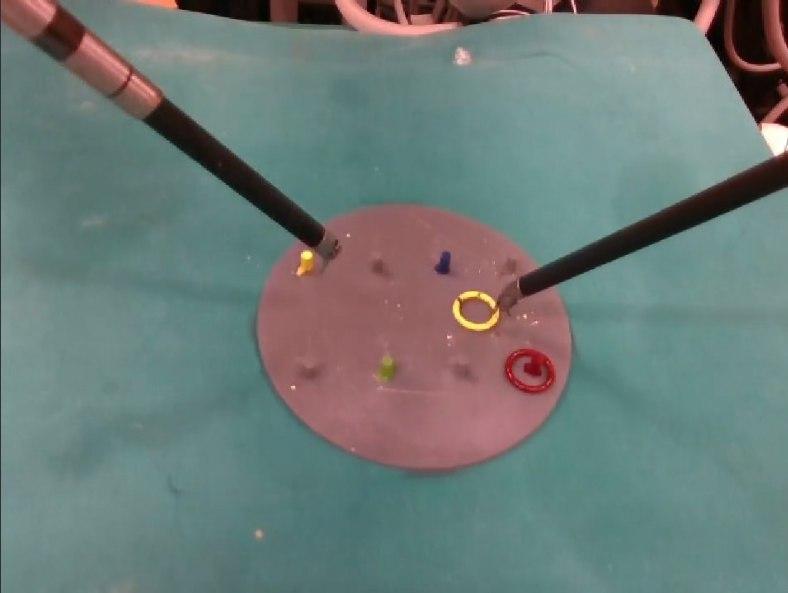}
\caption{move(psm1,ring,yellow,9)\label{fig:exec5}}
\end{subfigure}
\begin{subfigure}{0.22\textwidth}
\includegraphics[width=\textwidth]{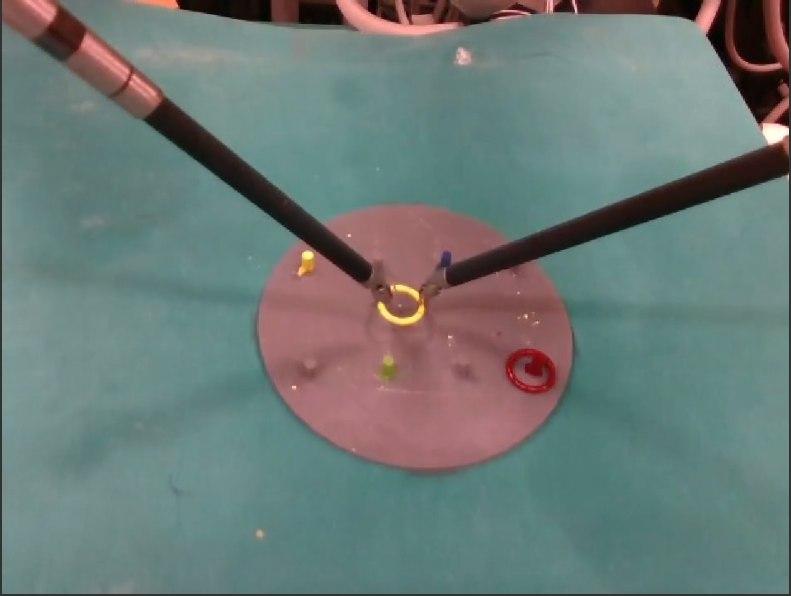}
\caption{move(psm1,center,11)\label{fig:exec6}}
\end{subfigure}
\begin{subfigure}{0.22\textwidth}
\includegraphics[width=\textwidth]{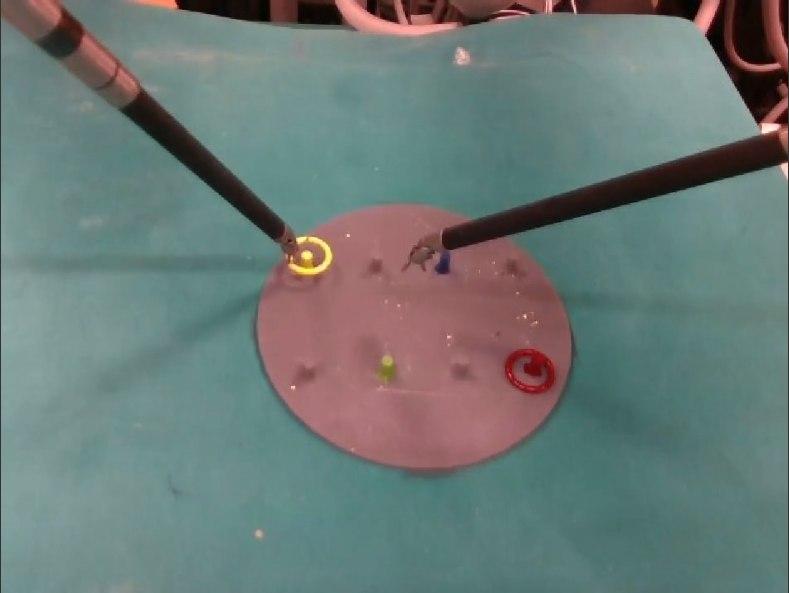}
\caption{move(psm2,peg,yellow,14)\label{fig:exec7}}
\end{subfigure}
\caption{Example execution of scenario \ref{fig:test_A}, with grounded actions for each state (\texttt{grasp, release} actions are omitted for simplicity). The initial plan is generated in a) after the grounding of the information from the SA module: \texttt{reachable(psm1,ring,red), reachable(psm1,ring,yellow), reachable(psm1,peg,red), reachable(psm1,peg,blue), reachable(psm2,peg,yellow), reachable(psm2,peg,green), on(ring,red,peg,grey)}. e) shows the re-planning of action \texttt{move(psm1,ring,yellow)} when \textit{ring\_fallen} is received from SA module.}
\label{fig:execution}
\end{figure*}

We validate our framework in three different scenarios, shown in the linked video and in Figure \ref{fig:tests}. In scenario \ref{fig:test_A} we show the main different versions of the peg-and-ring task: extraction is needed for the red ring which is placed on a grey peg, and transfer between arms is needed for the yellow ring. In spite of the calibration accuracy, the small size of the setup and light conditions sometimes originate vision errors. In this scenario, the reasoner is also able to re-plan when the first grasping of the yellow ring fails. In the video, we also show a similar scenario with the blue and red rings, where the system is able to recover from a different failure condition when transfer fails. In scenario \ref{fig:test_A}, we exploit preference reasoning in ASP to perform optimization and take the closest ring (red) first, using \texttt{distance} variable defined in Section \ref{sec:methods}.
In scenario \ref{fig:test_B}, colored pegs are occupied, hence a ring must be brought to a grey peg before starting the task. This operation is not encoded in the ASP program, but the ASP solver autonomously finds this solution to reach the final goal in the shortest time. Moreover, even if the SA module identifies the green ring, the robot does not operate on it since it is already placed on its peg.
Finally, in scenario \ref{fig:test_C} we test the simultaneous execution of the two arms to complete the task faster than standard execution by surgeons, using ASP aggregates as described in Section \ref{sec:methods}. Figure \ref{fig:execution} shows an example of the working of the ASP planner and its integration with the SA module when failure occurs (\ref{fig:exec5}), with reference to scenario \ref{fig:test_A}. In the video, we also show the grounded state variables in each discrete timestep, as identified by the SA module.

In Table \ref{table_example}, we show the task planning times for the tested scenarios. We also show the planning time for the standard scenario with all rings in the scene, to be transferred between arms. This is the worst-case scenario, since more actions are needed to reach the goal. The results prove the real-time capabilities of our task planner (1.78 s in worst-case scenario). We notice that optimization increases the planning time.

\begin{table}[h]
\caption{Planning time for the ASP task planner in the tested scenario and in the worst-case scenario (complete) with all four rings to be transferred between arms.}
\label{table_example}
\begin{center}
\begin{tabular}{|c||c|}
\hline
 & Planning time [s]\\
\hline
Scenario \ref{fig:test_A} (optimization) & 0.108 (0.424)\\
Scenario \ref{fig:test_B} & 0.133\\
Scenario \ref{fig:test_C} & 0.113\\
Complete (optimization) & 1.780 (8.636)\\
\hline
\end{tabular}
\end{center}
\end{table}

\section{CONCLUSIONS} \label{sec:conclusions}
In this paper, we have presented a framework for the autonomous execution of surgical tasks. To the best of our knowledge, this is the first framework which addresses the problems of failure recovery, explainable plan generation and situation awareness in the surgical scenario, which are required features for the acceptability of an autonomous surgical system. We have focused on a more complex version of the peg-and-ring task, a standard in the training program for surgeons, with the da Vinci surgical robot. This task requires the coordination of multiple actions in a non pre-defined way, depending on sensory information. Hence, an ASP-based task reasoner is implemented to find the fastest set of actions towards the goal, respecting a set of constraints. Motion trajectories are learned from tele-operated executions by users with different expertise, using the DMP framework to replicate human dexterity.

In future research we will test the framework on more surgically relevant operations. The automatic learning of ASP specifications from observed executions of surgical tasks will be investigated, in order to enrich the prior surgical knowledge. Higher-level surgical ontologies will be added to this framework to provide additional background knowledge which cannot be represented in the ASP planner.




\bibliographystyle{IEEEtran}
\bibliography{IEEEabrv,biblio.bib}
\end{document}